\documentclass[letterpaper, 10 pt, conference]{ieeeconf}  % Comment this line out if you need a4paper

\IEEEoverridecommandlockouts                              % This command is only needed if 
\usepackage{times}
\usepackage{graphicx} % in the preamble
\usepackage{cite}
\usepackage{multicol}
\usepackage[bookmarks=true]{hyperref}
\usepackage{caption}
\usepackage{amsmath}
\usepackage{cuted} % for full-width figure trick
\usepackage{booktabs, array} % Add this to your preamble

\usepackage[linesnumbered,ruled,vlined]{algorithm2e}

\usepackage{adjustbox}
\usepackage{multirow}
\usepackage{balance}
\usepackage{algorithmic}
\usepackage{amssymb} 

\usepackage{tikz}
\usepackage{pgfplots}
\usepgfplotslibrary{groupplots}
\usepackage{subcaption}
\pgfplotsset{compat=1.18}

\title{\LARGE \bf
TOPO-Bench: An Open-Source Topological Mapping Evaluation Framework with Quantifiable Perceptual Aliasing}

\author{
Jiaming Wang$^{*}$, 
Jizhuo Chen$^{*}$, 
Diwen Liu$^{*}$, Jiaxuan Da,\\
Jiamo Hu, Zhiwei Xue, Linh K\"astner, Harold Soh%
\thanks{$^{*}$Equal contribution.}%
\thanks{Emails: jiaming@comp.nus.edu.sg, \{e0905370, e0774920\}@u.nus.edu, harold@comp.nus.edu.sg.}
}

\begin{document}

% First-page publication note; does not change the text layout.
\AddToHookNext{shipout/foreground}{%
  \put(0,-35){\makebox[\paperwidth]{\small\itshape Published at ICRA 2026}}%
}

\maketitle
\thispagestyle{empty}
\pagestyle{empty}

%%%%%%%%%%%%%%%%%%%%%%%%%%%%%%%%%%%%%%%%%%%%%%%%%%%%%%%%%%%%%%%%%%%%%%%%%%%%%%%%

\begin{abstract}
Topological mapping offers a compact and robust representation for navigation, but progress in the field is hindered by the lack of standardized evaluation metrics, datasets, and protocols. Existing systems are assessed using different environments and criteria, preventing fair and reproducible comparisons. Moreover, a key challenge—perceptual aliasing—remains under-quantified, despite its strong influence on system performance. We address these gaps by (i) formalizing \emph{topological consistency} as the fundamental property of topological maps and showing that localization accuracy provides an efficient and interpretable surrogate metric, and (ii) proposing the first quantitative measure of dataset ambiguity to enable fair comparisons across environments. To support this protocol, we curate a diverse benchmark dataset with calibrated ambiguity levels, implement and release deep-learned baseline systems, and evaluate them alongside classical methods. Our experiments and analysis yield new insights into the limitations of current approaches under perceptual aliasing. All datasets, baselines, and evaluation tools are fully open-sourced to foster consistent and reproducible research in topological mapping.
\end{abstract}

\section{Introduction}

Topological maps represent the environment as a graph, where nodes correspond to discrete places and edges capture navigable connections between them. This discretization of the continuous world provides a compact and efficient representation, since only salient locations are stored and dense metric reconstructions are unnecessary. As a result, topological maps are far less computationally demanding than metric maps. Their efficiency is particularly relevant in the context of emerging mapless navigation systems~\cite{wang2025genie, sridhar2024nomad}, which operate using sparse subgoals that can be obtained through graph search on a topological map without relying on a full metric reconstruction.

Topological navigation is not only a useful abstraction in robotics but also a behavior observed in many animal species, including humans. In everyday movement, we rarely think in terms of exact coordinates; instead, we rely on abstract notions of distance and spatial relationships, recognizing where we are without requiring fine-grained metric accuracy. As argued by Brooks~\cite{brooks1991intelligence}, topological representations naturally help systems cope with uncertainty in mobile robot navigation. By replacing explicit metric information with notions of proximity and ordering, they avoid the accumulation of dead-reckoning errors that often undermine purely geometric approaches.

However, a persistent challenge in topological mapping is \emph{perceptual aliasing}~\cite{boal2014topological}, which is the risk that two physically distinct locations produce observations that appear identical to the robot's sensors. Perceptual aliasing can lead to incorrect loop closures and navigation failures, making its mitigation a central research question.

Despite the large number of topological mapping systems proposed in the literature~\cite{milford2008mapping, cummins2009highly, savinov2018semi}, there is currently no widely accepted or standardized benchmarking protocol, dataset, or evaluation metric. Existing works~\cite{glover2010fab, maddern2012cat, xu2020probabilistic, suomela2024placenav, ali2024boq, meng2020scaling} use different datasets, metrics, and environmental conditions, making direct comparisons between methods difficult and hindering systematic progress.

In this paper, we address this gap by proposing a principled evaluation protocol for topological mapping systems. Our metrics are derived from the core property of \emph{topological consistency} (Section~\ref{sec:topological-consistency}): if two nodes in the graph are connected by an edge, they should correspond to physically proximate locations along a traversable route in the real world, and conversely, such proximate locations should be connected. We then show that \emph{localization accuracy} (Section~\ref{sec:localization-success}) provides an efficient and interpretable surrogate metric for topological consistency.

We further highlight that perceptual aliasing is not only a system-level challenge but also a dataset property: the number and difficulty of ambiguous observations in the test environment strongly influence performance. Without controlling for this \emph{ambiguity factor}, comparisons across datasets are unreliable (Section~\ref{sec:results}). We propose a quantitative method to measure dataset ambiguity, enabling more meaningful evaluations (Section~\ref{sec:ambiguity-quantification}).

Following this protocol, we curate a diverse set of scenes and design targeted test cases. We evaluate several baseline systems, spanning common strategies for  topological localization, on our curated datasets. Our results show that methods relying solely on topological structure struggle in the presence of perceptual aliasing, highlighting the need for additional cues beyond topology to achieve reliable mapping and localization (Section~\ref{sec:results}).

In summary, our contributions are:
\begin{enumerate}
    \item We systematically define metrics for evaluating topological mapping systems and introduce the first framework, to our knowledge, for quantifying dataset ambiguity in this context.
    \item We curate and release a diverse, publicly available dataset and evaluation toolkit to support consistent and reproducible benchmarking.
    \item We implement and open-source baseline systems using deep-learned models, and evaluate them alongside classical methods; our experiments and analysis yield insights into the limitations of current approaches and factors affecting robustness under perceptual aliasing.
\end{enumerate}

\section{Background and Related Work}
Broadly speaking, topological maps can be categorized into two types: those constructed \emph{with} a metric SLAM map and those constructed \emph{without} one (SLAM-free). The former relies on a metrically consistent representation (e.g., occupancy grid~\cite{wang2024polo}, TSDF/ESDF volume~\cite{blochliger2018topomap, oleynikova2018sparse}, or pose graph) and builds a sparse topological abstraction on top of it to enable efficient planning and high-level reasoning. The latter constructs the topological map directly from sensory data, without requiring a globally consistent metric map. In this paper, our focus is \textbf{SLAM-free topological mapping systems without a metric consistent map from SLAM}.

\subsection{SLAM-based Topological Maps}
These methods start from a metric SLAM output and derive a sparse topological graph for planning. For example, Bl{\"o}chliger et al.~\cite{blochliger2018topomap} convert a TSDF map into convex free-space clusters connected by adjacency edges, enabling A* search with near–RRT* path quality but much faster computation. Oleynikova et al.~\cite{oleynikova2018sparse} extract a Generalized Voronoi Diagram from an ESDF and prune it into a thin 3D skeleton for robust graph generation in MAV planning. Konolige et al.~\cite{konolige2011hybrid} overlay a navigation graph on a SLAM pose graph while maintaining only local occupancy grids for execution. Friedman et al.~\cite{friedman2007vrf} and Beeson et al.~\cite{beeson2005evg} extract Voronoi skeletons from occupancy grids and segment them into meaningful places such as rooms, doorways, and corridors. While these approaches produce reliable connectivity graphs and large planning speedups, they inherit the computational and maintenance costs of building a metrically consistent map, which can become a bottleneck in large or dynamic environments.

\subsection{SLAM-free Topological Maps}
\label{sec:slam-free-topological-maps}

These methods construct and maintain the topological map directly from observations, avoiding a globally consistent metric map. Milford et al.~\cite{milford2008mapping} and Glover et al.~\cite{glover2010fab} use bio-inspired pose cell dynamics to filter place-recognition proposals from visual template matching. Cummins and Newman~\cite{cummins2009highly} and Maddern et al.~\cite{maddern2012cat} adopt a Bayesian filter where the observation model is learned via a Chow--Liu tree to capture word correlations, and the motion prior models topological distance. Learning-based methods such as Savinov et al.~\cite{savinov2018semi} train image-retrieval networks and use a sliding window to detect related frames, while Suomela et al.~\cite{suomela2024placenav} and Xu et al.~\cite{xu2020probabilistic} integrate Bayesian filtering for robust retrieval. However, the variety of datasets and evaluation metrics used makes direct comparison between methods difficult.

\begin{figure}[t!]
    \centering
    \includegraphics[width=0.98\linewidth]{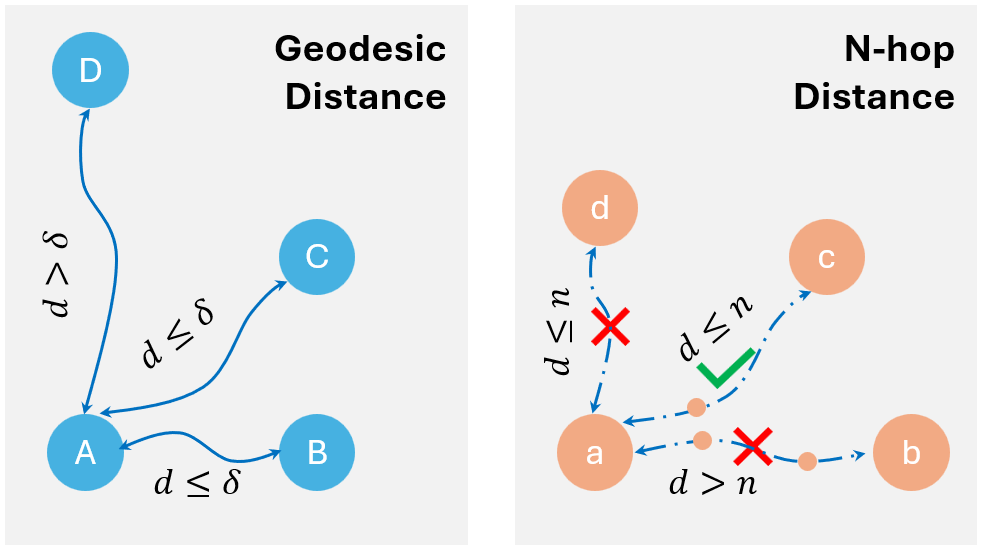}
    \caption{Illustration of topological consistency. Nodes $a$, $b$, $c$, and $d$ correspond to physical locations $A$, $B$, $C$, and $D$ in the environment. Location $A$ is route-close to $B$ and $C$, but far from $D$. In the constructed topological map, the path between $a$ and $d$ violates \emph{Edge Precision}, while the missing edge between $a$ and $b$ violates \emph{Edge Recall}. The connection between $a$ and $c$ correctly preserves topological consistency.}
    \label{fig:topo_consistency}
\end{figure}

\begin{figure*}[t!]
\centering

% ---------- Row 1: three images, each ~1/3 of textwidth ----------
\begin{minipage}{\textwidth}
  \centering
  \includegraphics[width=0.33\textwidth]{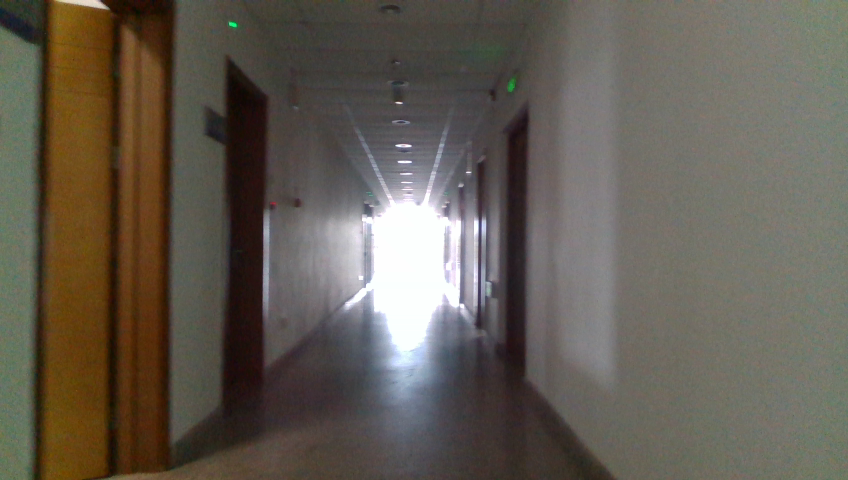}%
  \includegraphics[width=0.33\textwidth]{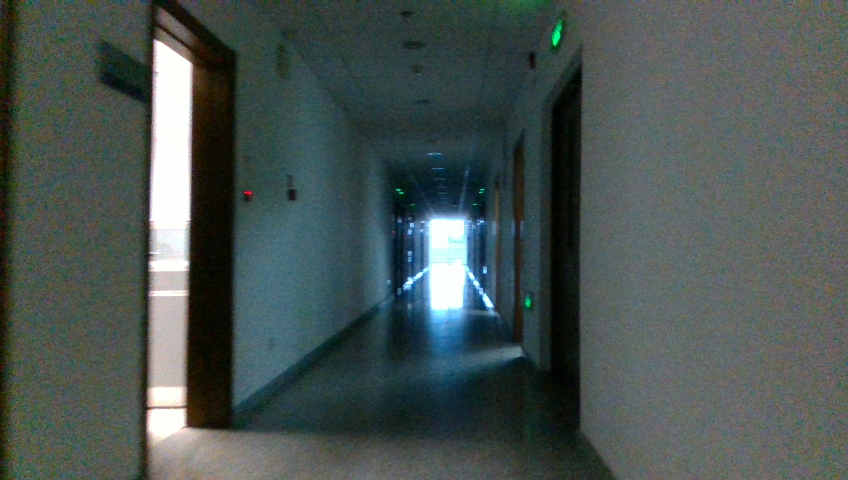}%
  \includegraphics[width=0.33\textwidth]{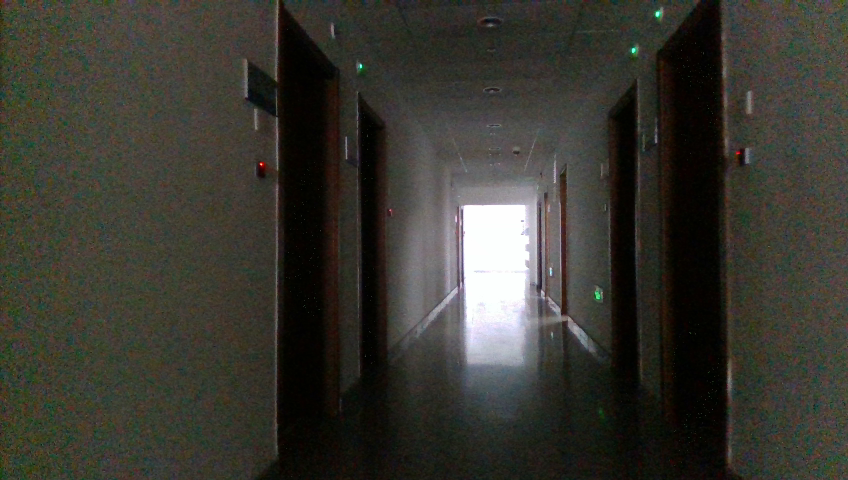}%
\end{minipage}

\vspace{0.8ex}

% ---------- Row 2: force equal height for all four images ----------
\newlength{\rowBheight}
\setlength{\rowBheight}{0.23\textwidth} % adjust this if needed
\begin{minipage}{\textwidth}
  \begin{minipage}{0.495\textwidth}
    \centering
    \includegraphics[width=0.495\textwidth,height=\rowBheight]{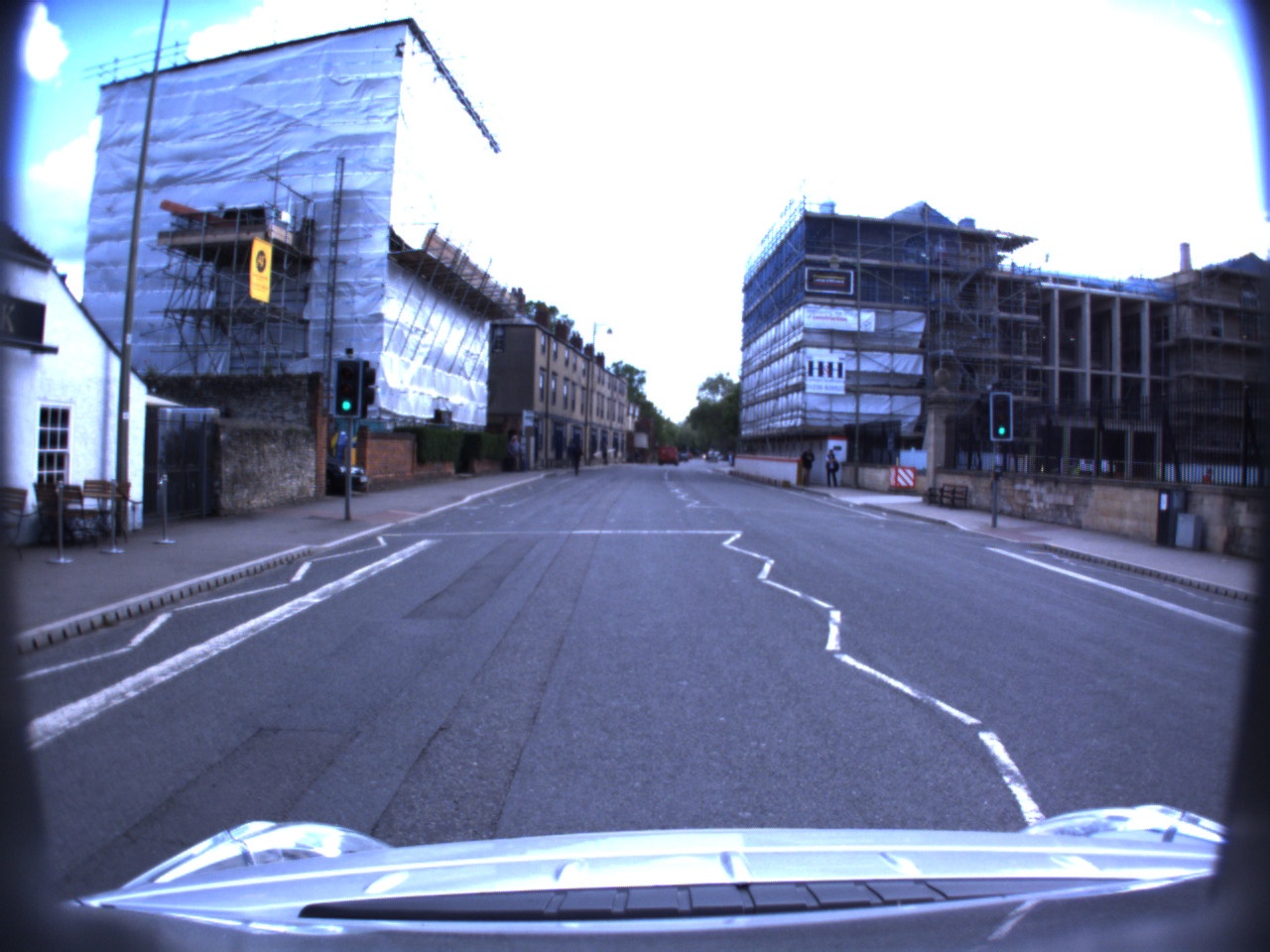}%
    \includegraphics[width=0.495\textwidth,height=\rowBheight]{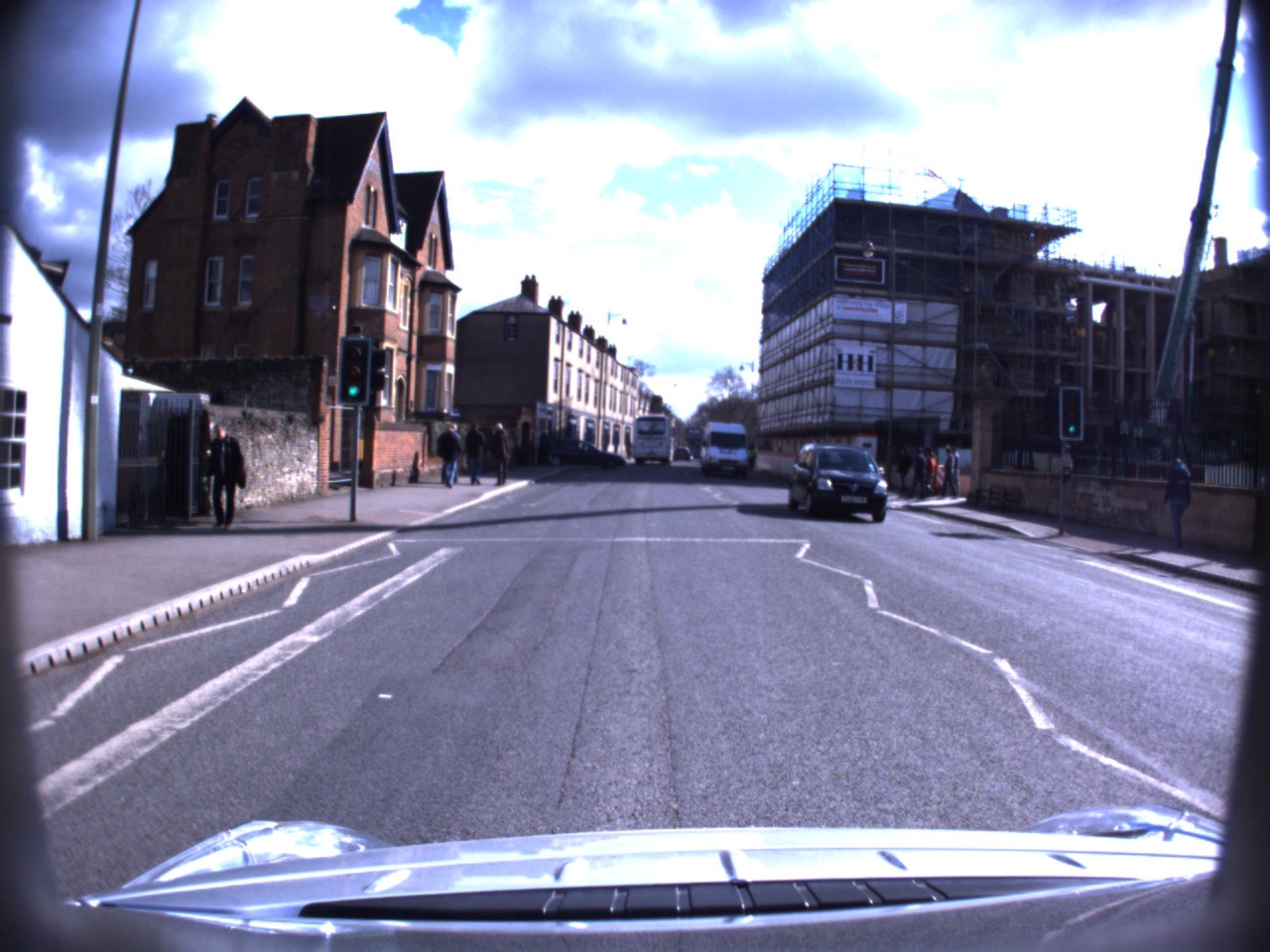}%
  \end{minipage}\hfill
  \begin{minipage}{0.495\textwidth}
    \centering
    \includegraphics[width=0.495\textwidth,height=\rowBheight]{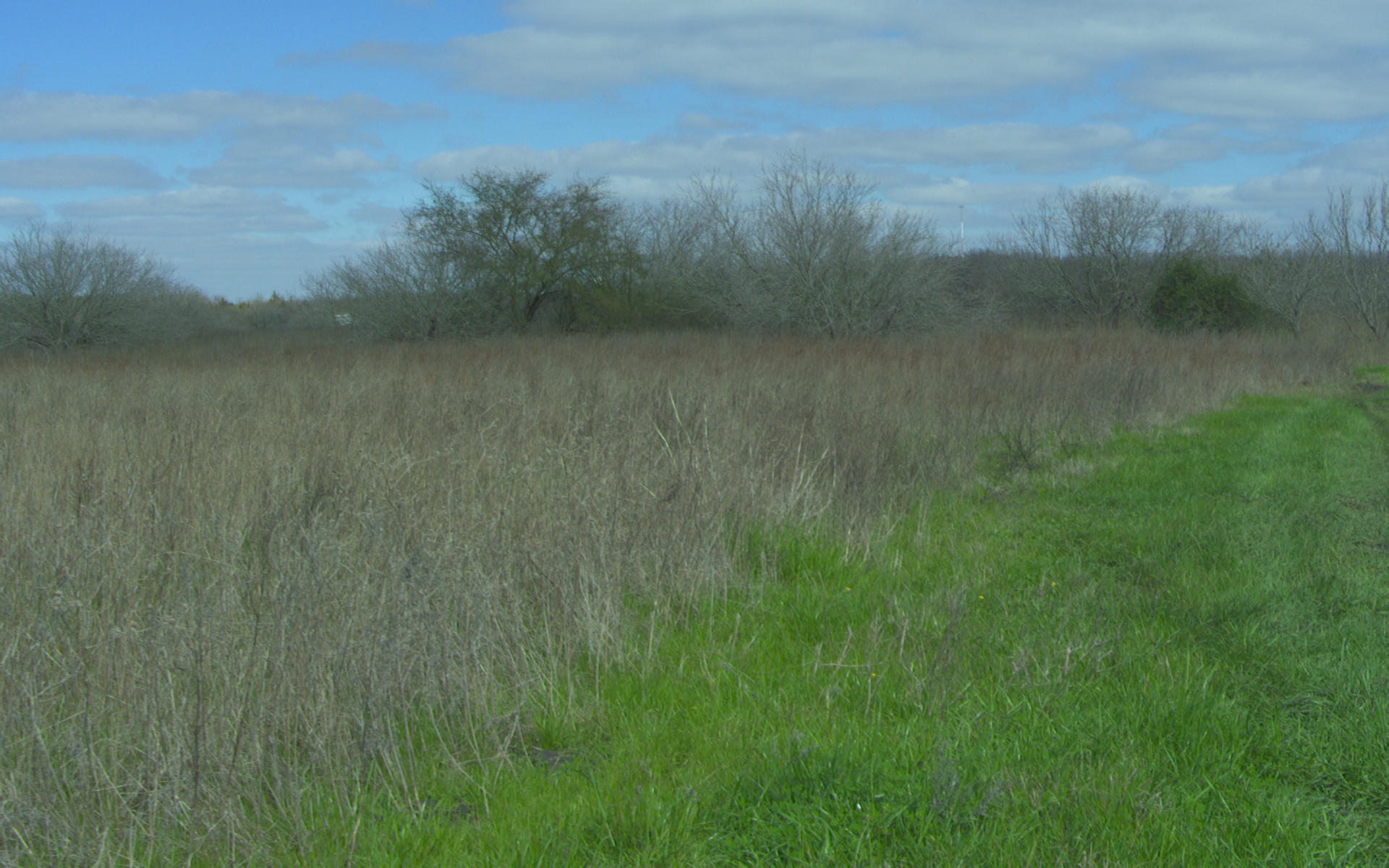}%
    \includegraphics[width=0.495\textwidth,height=\rowBheight]{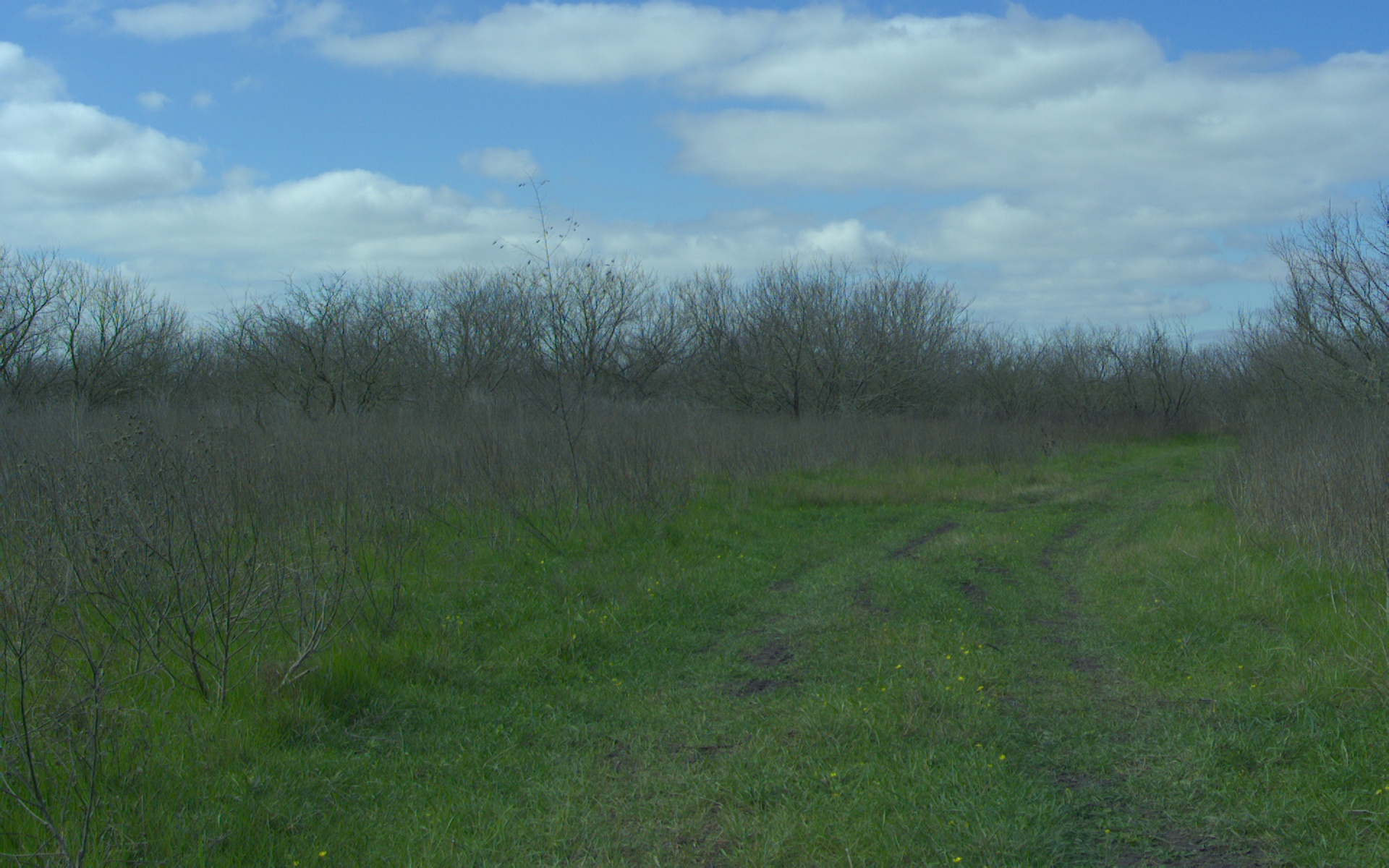}%
  \end{minipage}
\end{minipage}

\caption{Examples of map, test, and ambiguous cases across datasets. 
Top row: A+P case from OpenLORIS~\cite{shi2019openlorisscene}, where the test differs from the map due to a time-of-day change, and the ambiguous (false-positive) image appears visually close to the target. 
Bottom left: P.O. case from RobotCar~\cite{RobotCarDatasetIJRR}, showing a map–test pair of the same location with the building façade temporarily covered during construction. 
Bottom right: A.O. case from RELLIS-3D~\cite{jiang2020rellis3d}, where the ambiguous resembles the test except for differences in vegetation type. See Sec.~\ref{sec:ambiguity-quantification} for case definitions.}

\label{fig:test_cases}
\end{figure*}

\section{Evaluation on Topological Systems}
\label{sec:evaluation-on-topological-systems}

\subsection{Preliminaries: Topological Mapping}
\label{sec:preliminaries}
In contrast to SLAM systems, where each location is assigned a global metric pose and evaluated using distances in Euclidean or $SE(3)$ space, topological maps do not aim to preserve global metric consistency. Instead, they represent the environment as a connected graph
$
\mathcal{G} = (\mathcal{V}, \mathcal{E}),
$
with nodes $\mathcal{V}$ denoting distinct places or observations and edges $\mathcal{E}$ denoting navigable connections.

We adopt a generic framework that reflects the core procedure of many topological mapping systems. At each timestep $t$, given a new observation $z_t$, the system follows an \emph{update policy} $P$ that governs how nodes and edges are added:
\begin{itemize}
    \item The agent first attempts to \emph{localize} in the existing graph $\mathcal{G}$. If a node $v_j \in \mathcal{V}$ is successfully retrieved (above a similarity threshold $\tau$), an edge $(v_i, v_j)$ is created between the current node $v_i$ and the matched node $v_j$.
    \item If no suitable match is found, a new node $v_{i+1}$ is created when $z_t$ differs sufficiently from prior observations and/or the agent has moved beyond a spatial threshold from the last node. This new node is then connected to $v_i$ with an edge $(v_i, v_{i+1})$.
\end{itemize}

In this work, we consider $P$ to be an \emph{edge-length regular policy}, where each new edge has a bounded geodesic route length proportional to the median edge length $\mu_e$. This mild assumption reflects common practice, as edges are typically added only between consecutive nodes once the robot has moved beyond a threshold. Focusing on edge-length regular policies remains general, since shortcut edges or other connections can always be added later through post-processing.

Based on this policy, we define an \emph{edge creation opportunity} as a pair of nodes $(u,v)$ that $P$ could have connected at the evaluation scale $d$. The set of such pairs, denoted $\Omega_P(d)$, specifies the policy-eligible connections that should be considered when evaluating recall.

\subsection{Topological Consistency}
\label{sec:topological-consistency}

For a topological map to be meaningful and useful for representing the environment and enabling navigation planning, it should satisfy two properties:

\begin{enumerate}
    \item \textbf{Edge Precision:} If two nodes in the topological graph are within $n$ hops of each other, then their corresponding locations in the environment should be within a geodesic route distance threshold $d$.
    \item \textbf{Edge Recall (policy-conditioned):} If two locations in the environment are within a geodesic route distance threshold $d$, and their corresponding nodes form an edge creation opportunity under the update policy $P$ (i.e., $(u,v)\in\Omega_P(d)$), then those nodes should be connected within $n$ hops in the graph.
\end{enumerate}

This concept is illustrated in Fig.~\ref{fig:topo_consistency}.  
We use \emph{route distance} (the shortest traversable geodesic in the environment) rather than Euclidean distance, since topological maps are intended to capture navigable paths where direct straight-line connections may not exist. In practice, ground-truth route distances are rarely available, so we approximate them using traversal distances obtained from the robot’s odometry. During dataset curation, we manually verified that no repeated zig-zag motions were present, ensuring that traversal distances do not overestimate the true geodesic distance.

Different mapping systems produce edges of varying lengths, so fixed thresholds $(n,d)$ may bias evaluation. We instead define a characteristic edge length using the median, which robustly captures a “typical” edge:  
\begin{equation}
\label{eq:median_edge_length}
\mu_{e} \ :=\ \mathrm{median}\{\mathrm{dist}_{R}(u,v) \mid (u,v) \in \mathcal{E}\},
\end{equation}
where $\mathrm{dist}_{R}(u,v)$ is the geodesic route distance between the locations of nodes $u$ and $v$.

Given a physical route distance threshold $d$, we adapt the hop threshold as  
$n = \max\!\left(1, \left\lfloor \tfrac{\epsilon d}{\mu_e} \right\rfloor \right)$,  
where $\epsilon > 0$ is a tolerance factor and the floor ensures $n \geq 1$.  
Evaluation is thus based on the physical route distance $d$, as navigation requires maps to support localization and connectivity within a bounded physical distance. Since $d$ is usually larger than the typical edge length $\mu_e$, $n$ often exceeds 1.

With this definition, the two properties can be formalized as:
\begin{align} &\forall u, v \in \mathcal{V}, \quad \mathrm{dist}_{G}(u, v) \leq n \ \Rightarrow\ \mathrm{dist}_{R}(u, v) \leq d, \label{eq:topo_consistency_precision} \\[3pt] &\forall (u, v) \in \Omega_P(d), \quad \mathrm{dist}_{R}(u, v) \leq d \ \Rightarrow\ \mathrm{dist}_{G}(u, v) \leq n, \label{eq:topo_consistency_recall} \end{align}

where $\Omega_P(d)$ is the set of policy-eligible edge opportunities (Section~\ref{sec:preliminaries}), $\mathrm{dist}_{G}(\cdot,\cdot)$ is shortest-path hop distance in the graph, and $\mathrm{dist}_{R}(\cdot,\cdot)$ is geodesic route distance in the environment. 

Equation~\eqref{eq:topo_consistency_precision} defines \emph{Edge Precision}: nodes that are close in the graph (within $n$ hops) should also be geodesically close in the environment (within $d$).  
Equation~\eqref{eq:topo_consistency_recall} defines \emph{Policy-conditioned Edge Recall}: geodesically close locations should be connected within $n$ hops, but only if the update policy $P$ makes such a connection feasible.  

This policy-aware definition avoids penalizing systems for missing edges they would never create (e.g., between two nearby but non-consecutive nodes), while still enforcing that feasible edge opportunities preserve navigability.

\subsection{Localization Accuracy}
\label{sec:localization-success}

While edge precision and recall (Section~\ref{sec:topological-consistency}) measure structural consistency, a topological map must also support reliable \emph{localization} for safe navigation. That is, given the current sensor observation, the system should correctly identify the robot’s current node in the existing graph. 

Formally, let $\hat{v}_t$ be the predicted node at time $t$ and $v^*_t$ the ground-truth node. Localization is considered successful if the geodesic route distance between them is within a physical threshold $d$:
\begin{equation}
\forall t,\quad \mathrm{dist}_{R}\!\left(\hat{v}_t, v^*_t\right) \le d.
\label{eq:localization_success_e}
\end{equation}
If the test observation lies in a novel region with no corresponding ground-truth node, then success is defined by correct rejection: accuracy is $1$ if the system abstains from matching, and $0$ otherwise. 

We argue that if the system makes \emph{correct localization decisions} at each step, then Edge Precision and policy-conditioned Edge Recall are preserved as the map grows, making localization accuracy a faithful surrogate for topological consistency. We provide a brief sketch proof by induction below.

Since the update policy $P$ is \emph{edge-length regular} (Sec.~\ref{sec:preliminaries}), each new edge has geodesic route length at most $\kappa\,\mu_e$ for some $\kappa\!\ge\!1$. Choose the hop budget $n=\max(1,\lfloor \epsilon d/\mu_e\rfloor)$ with $\epsilon\!\le\!1/\kappa$, so that any $n$-hop path has geodesic length $\le n\,\kappa\,\mu_e \le d$.

Suppose at time $t$ the map $\mathcal{G}_t$ already satisfies Edge Precision and policy-conditioned Edge Recall at scale $(n,d)$. After processing $z_{t+1}$ there are two cases:

\emph{(i) Correct accept (revisit).} The system links the current node $v_i$ to the correctly retrieved node $v_j$ with a new edge. For \textbf{Edge Precision}, any pair within $n$ hops in $\mathcal{G}_{t+1}$ either (a) was already within $n$ hops in $\mathcal{G}_t$ and thus satisfies $\mathrm{dist}_R\!\le d$ by induction, or (b) uses the new edge once; in this case the path length is at most $n\,\kappa\,\mu_e \le d$, so $\mathrm{dist}_R\!\le d$. For \textbf{policy-conditioned Edge Recall}, all previously eligible pairs remain connected within $n$ hops, and any new eligible pair (in particular $(v_i,v_j)$) is directly connected, i.e., $\mathrm{dist}_G(v_i,v_j)=1\le n$.

\emph{(ii) Correct reject (novel).} The system creates a new node $u$ and a consecutive edge $(v_i,u)$. For \textbf{Edge Precision}, any $n$-hop path either existed before (hence $\mathrm{dist}_R\!\le d$) or uses the new edge once, which again yields $\mathrm{dist}_R\!\le d$. For \textbf{policy-conditioned Edge Recall}, the only new eligible pair is $(v_i,u)$, which is connected in one hop, while all previously eligible pairs retain their $n$-hop connectivity.

Thus, under correct localization decisions, both properties are invariants: if they hold at time $t$, they also hold at $t+1$. Since the base map (empty or a single node) trivially satisfies them, they hold at all times. This establishes localization accuracy as a concise, interpretable, and faithful surrogate for topological consistency. In summary, localization accuracy is directly linked to both navigation success and successful map construction.

\begin{figure}[t]
    \centering
    \begin{subfigure}[t]{0.32\linewidth}
        \centering
        \includegraphics[width=\linewidth]{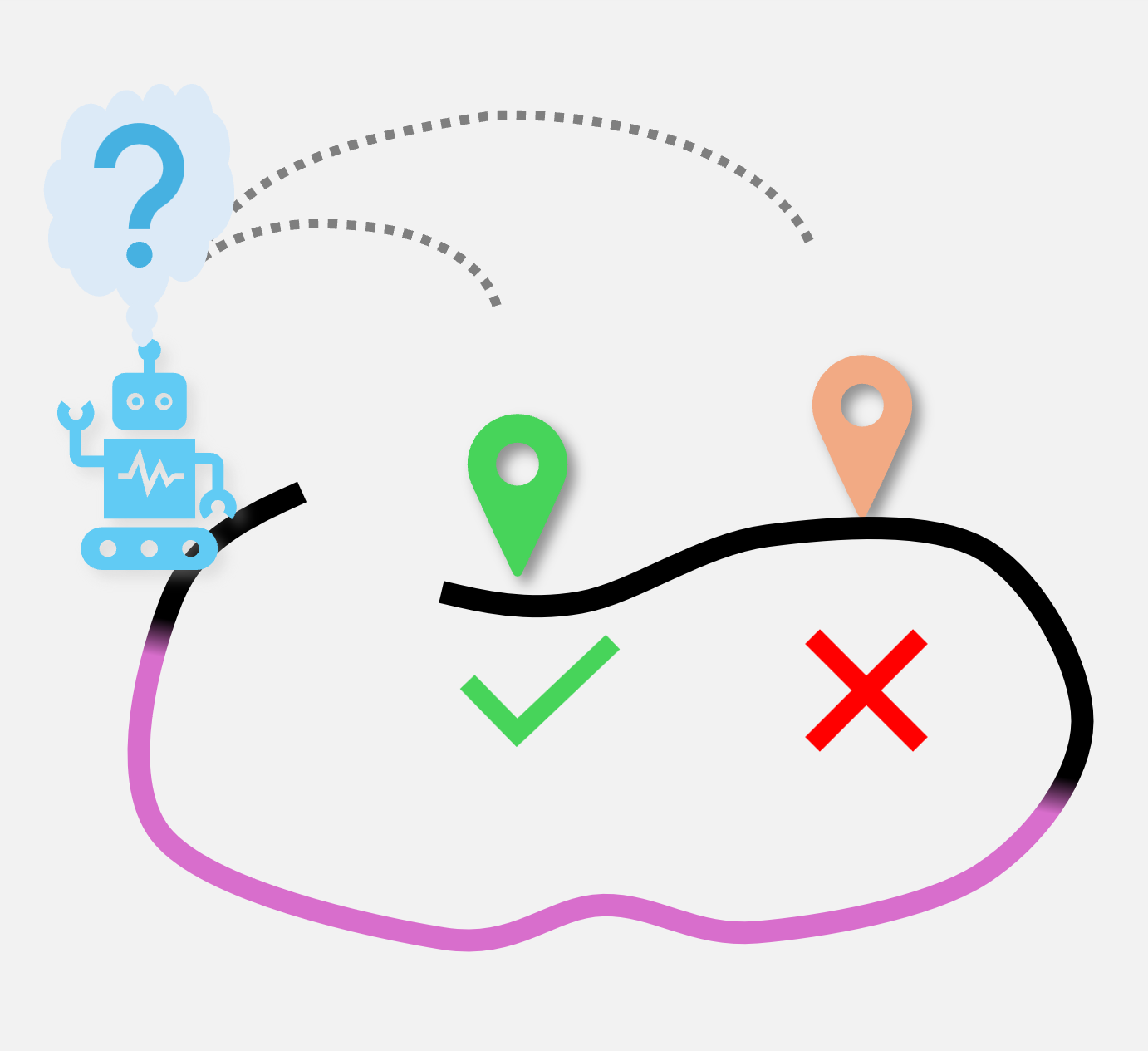}
        \caption{A+P}
        \label{fig:test_ap}
    \end{subfigure}
    \hfill
    \begin{subfigure}[t]{0.32\linewidth}
        \centering
        \includegraphics[width=\linewidth]{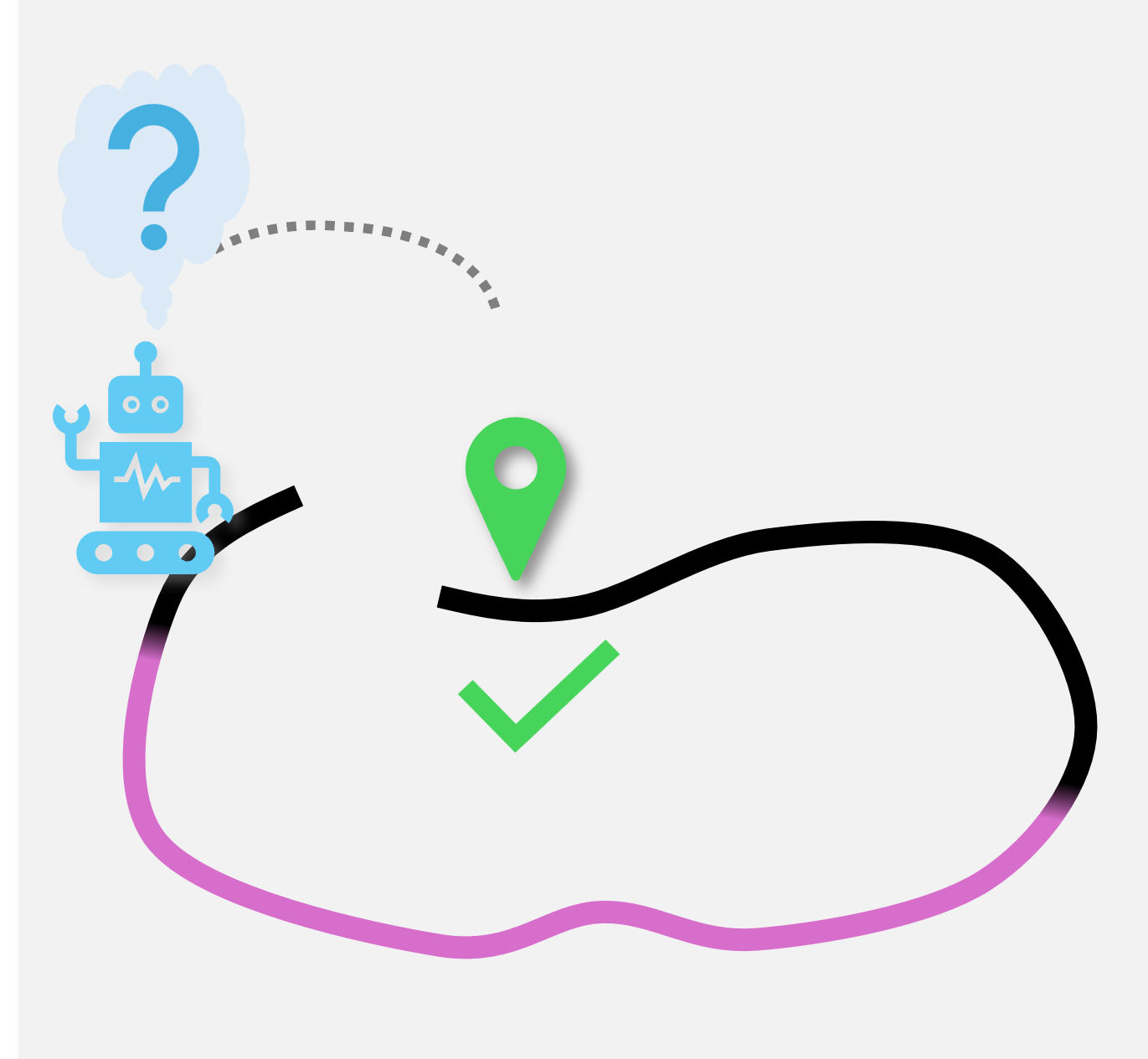}
        \caption{P.O.}
        \label{fig:test_po}
    \end{subfigure}
    \hfill
    \begin{subfigure}[t]{0.32\linewidth}
        \centering
        \includegraphics[width=\linewidth]{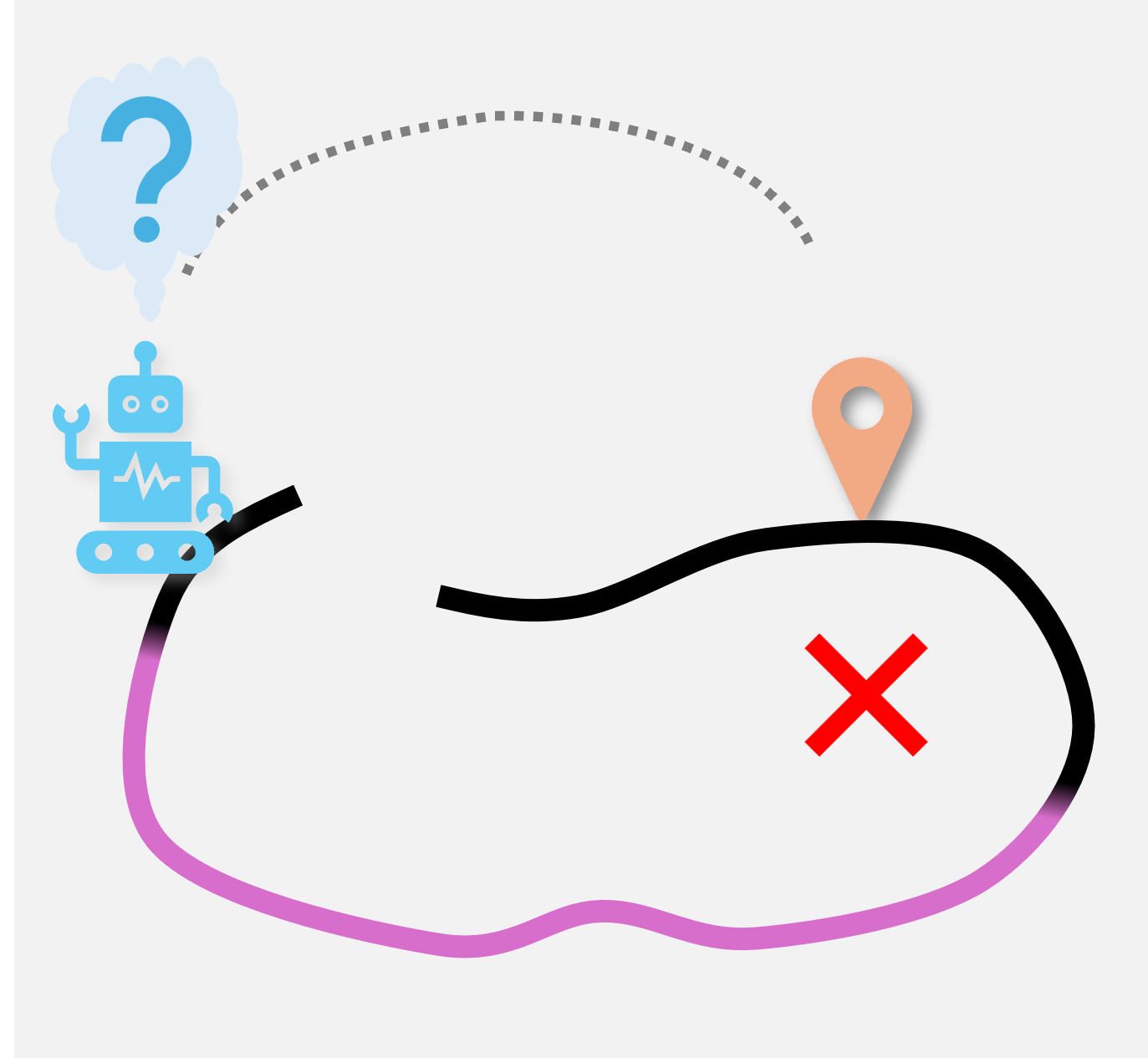}
        \caption{A.O.}
        \label{fig:test_ao}
    \end{subfigure}
    \caption{Examples of evaluation scenarios with quantified ambiguity. 
    (a) \textbf{Ambiguous + Positive (A+P):} The test sequence revisits a mapped region, but at least one distractor subsequence appears nearly as similar as the true match, making localization uncertain. 
    (b) \textbf{Positive Only (P.O.):} The test sequence revisits a mapped region and the true subsequence clearly dominates all distractors, resulting in unambiguous localization. 
    (c) \textbf{Ambiguous Only (A.O.):} The test sequence lies in a novel, unmapped region, yet one or more mapped subsequences spuriously appear visually similar, producing false matches. 
    Omitting the intermediate trajectory (shown in magenta) yields the kidnapped-robot variant; otherwise, the scenario corresponds to a classical loop closure.}
    \label{fig:evaluation_scenarios}
    \vspace{-1em}
\end{figure}

\subsection{Ambiguity Quantification}
\label{sec:ambiguity-quantification}

Perceptual aliasing, where distinct locations produce highly similar sensor observations, is a core challenge in topological mapping \cite{boal2014topological}. It can cause the map to mistakenly add edges between unrelated places, leading to structural corruption. Such ambiguity arises across diverse environments, from uniform office corridors and near-identical urban intersections to repetitive patterns in natural trails (Fig.~\ref{fig:test_cases}). Yet most prior work evaluates mapping performance without explicitly measuring or controlling for this phenomenon, resulting in inconsistent and potentially misleading comparisons across systems and environments. This motivates our explicit quantification of ambiguity as a central element of evaluation.

To incorporate ambiguity-aware evaluation, we adopt the following protocol. First, the system is provided with a \emph{mapping sequence} $\mathcal{S}_m = \{ z_m^j \}_{j=1}^{N}$ followed by a short \emph{test sequence} $\mathcal{S}_t = \{ z_t^i \}_{i=1}^{L}$. The test sequence may revisit previously mapped regions or explore an entirely new place. Second, we evaluate localization accuracy (Section~\ref{sec:localization-success}) using the final frame of $\mathcal{S}_t$. Here, $z_m^j$ and $z_t^i$ denote individual sensor observations (e.g., images) in the mapping and test phases, respectively. Using ground truth (e.g., GPS or motion capture), each test frame $z_t^i$ is aligned to its corresponding mapping frame via an index function $\pi(i) \in \{1, \dots, N\}$, such that $z_m^{\pi(i)}$ is the true spatial match. If the test sequence corresponds to a novel place, we set $\pi(i)=\varnothing$. This setup accommodates both loop closure and kidnapped-robot scenarios (Fig.~\ref{fig:evaluation_scenarios}).

We define ambiguity at the level of short \emph{sequences} of frames. For a candidate subsequence of length $L$,  $\mathcal{S}_m^{[u]} = \{ z_m^{j_u}, \dots, z_m^{j_u+L-1} \}$,
we measure similarity as  
\[
\mathrm{Sim}(\mathcal{S}_t, \mathcal{S}_m^{[u]}) = \frac{1}{L} \sum_{i=1}^{L} \mathrm{sim}(z_t^i, z_m^{j_u + i - 1}),
\]  
where $\mathrm{sim}(\cdot,\cdot)$ denotes the similarity between two frames, provided by a visual place recognition (VPR) model~\cite{lowry2015visual}. A sequence pair is deemed visually similar if $\mathrm{Sim}(\cdot,\cdot) \ge \tau$, with $\tau$ a predefined threshold.

To validate the VPR model choice, we curated a balanced dataset of 200 image pairs spanning a wide range of appearance similarities. We evaluated 11 VPR methods reviewed in~\cite{berton2023eigenplaces}, and found a mean inter-method correlation of \textbf{0.83}. This high agreement suggests that ambiguity is an intrinsic property, largely independent of the specific method used. For dataset construction, we adopted BOQ~\cite{ali2024boq} due to its efficiency and robustness.

Under this definition, each test sequence falls into one of three categories (Fig.~\ref{fig:evaluation_scenarios}):

\textbf{Ambiguous + Positive (A+P):}
\[
\pi(\mathcal{S}_t) \neq \varnothing,\quad
\frac{\max_{u \neq \pi(\mathcal{S}_t)} \ \mathrm{Sim}(\mathcal{S}_t, \mathcal{S}_m^{[u]})}
     {\mathrm{Sim}(\mathcal{S}_t, \mathcal{S}_m^{[\pi(\mathcal{S}_t)]})} \ge \alpha.
\]
The sequence revisits a known region, but at least one distractor subsequence in the map appears almost as similar as the true match. Such cases often arise from scene changes in the test sequence (e.g., variations in lighting or time of day) that narrow the confidence gap between true and distractor matches.

\textbf{Positive Only (P.O.):}
\[
\pi(\mathcal{S}_t) \neq \varnothing,\quad
\frac{\max_{u \neq \pi(\mathcal{S}_t)} \ \mathrm{Sim}(\mathcal{S}_t, \mathcal{S}_m^{[u]})}
     {\mathrm{Sim}(\mathcal{S}_t, \mathcal{S}_m^{[\pi(\mathcal{S}_t)]})} < \alpha.
\]
The true match clearly dominates all distractors, making localization unambiguous.  

\textbf{Ambiguous Only (A.O.):}
\[
\pi(\mathcal{S}_t) = \varnothing,\quad
\max_{u} \ \mathrm{Sim}(\mathcal{S}_t, \mathcal{S}_m^{[u]}) \ge \tau.
\]
The test sequence lies in a novel region, but one or more mapped subsequences appear spuriously similar.  

\medskip
\noindent\textit{Discussion.} Our protocol improves on loop-closure precision/recall in two ways. First, it is \emph{ambiguity-aware}: by separating Positive-Only (P.O.), Ambiguous+Positive (A+P), and Ambiguous-Only (A.O.), it reveals failure modes that pairwise metrics conflate, such as disambiguation errors versus unsafe matches in novel regions. Second, it evaluates the robot’s \emph{query-time decision}, that is, whether to localize or abstain, rather than candidate-pair rates. This better reflects safe navigation needs, where a single wrong match can corrupt the map. In contrast, precision/recall may look strong even when top-1 mapping decisions are unreliable, since they summarize candidate pairs rather than end-to-end reliability.

Still, comparing systems by per-case accuracies is difficult. A method may excel on P.O. yet fail on A.O. To address this, we introduce a summary metric that balances performance across all case types.

\begin{table}[t]
\centering
\caption{Summary of datasets with environment counts, types (In./Out.), and difficulty-controlled test cases.}
\label{tab:dataset_stats_transposed}
\begin{tabular}{lccccccc}
\toprule
 & \cite{shi2019openlorisscene} & \cite{RobotCarDatasetIJRR} & \cite{fontana2013rawseeds} & \cite{habitat19iccv} & \cite{jiang2020rellis3d} & \cite{schmidt2025rover} & Total \\
\midrule
Env (No.) & 5 & 1 & 1 & 16 & 1 & 1 & 25 \\
Env (Type) & In. & Out. & In. & In. & Out. & Out. & -- \\
A+P & 30 & 0 & 16 & 0 & 0 & 5 & 51 \\
P.O. & 74 & 116 & 90 & 74 & 0 & 30 & 384\\
A.O. & 52 & 0 & 14 & 3 & 117 & 8 & 194 \\
\bottomrule
\end{tabular}
\end{table}

\subsection{Balanced Localization Accuracy}
\label{sec:balanced-localization}

The \emph{Balanced Localization Accuracy (BLA)} metric provides a unified measure of performance across the three case types defined in Section~\ref{sec:ambiguity-quantification}. It is computed as the geometric mean of localization accuracies:
\[
\mathrm{BLA} = \big(\mathcal{L}_{A+P}\,\mathcal{L}_{P.O.}\,\mathcal{L}_{A.O.}\big)^{1/3},
\]
where $\mathcal{L}_{\cdot}$ denotes accuracy on the respective case type. The geometric mean penalizes imbalance: a system that excels on one case type but fails on another will see its overall score reduced, analogous to how the $F$-score balances precision and recall.

To avoid degenerate cases when any accuracy is exactly zero, we apply \emph{Jeffreys prior smoothing}, a Bayesian adjustment with $\mathrm{Beta}(\tfrac{1}{2},\tfrac{1}{2})$ prior. Each empirical rate $r=k/n$ is replaced with $\tilde r = \tfrac{k+0.5}{n+1}$, which prevents collapse and has negligible effect when $n$ is large.

Each accuracy $\mathcal{L}_{\cdot}$ depends on the decision threshold $\tau$, which controls whether a retrieved candidate is accepted or rejected. Consequently, the BLA also varies with $\tau$. Typically, increasing $\tau$ improves rejection of ambiguous observations (raising $\mathcal{L}_{A.O.}$) but reduces acceptance of correct matches (lowering $\mathcal{L}_{P.O.}$ and $\mathcal{L}_{A+P}$).

In our experiment we report the accuracy $\mathcal{L}$ for each test case and the corresponding BLA, with the threshold $\tau$ chosen in two ways. First, $\mathcal{L}_{A.O.}@\rho$ denotes the operating point where $\mathcal{L}_{A.O.}(\hat\tau)\approx\rho$, within a tolerance of $\pm 3\%$ and preferring the conservative choice $\mathcal{L}_{A.O.}(\hat\tau)\ge\rho$, reflecting performance at a desired reliability level (e.g., $\rho=0.90$ or $0.99$). Second, $\mathrm{BLA}_{\max}$ is obtained by selecting the threshold $\hat\tau$ that maximizes BLA across all values, representing the best performance achievable under ideal tuning. In both cases, the threshold $\hat{\tau}$ is determined on a smaller validation set and then applied for evaluation on the full dataset, so the resulting $\mathcal{L}$ and BLA values may differ between the two stages.

\subsection{Dataset Collection}
\label{sec:dataset-collection}

To evaluate robustness across diverse environments, sensing setups, and robot embodiments, we compose a benchmark from datasets that provide time-synchronized observations and ground-truth trajectories (required for the correspondence function $\pi(\cdot)$ in Sec.~\ref{sec:ambiguity-quantification}). We include both indoor and outdoor scenes, and select datasets that naturally contain visual ambiguities through repeated traversals under varying conditions.

We consider the following datasets:
\begin{enumerate}
    \item \textbf{OpenLORIS}~\cite{shi2019openlorisscene}: Multiple traversals of everyday indoor environments with variations in activity, lighting, and object placement causing perceptual aliasing.
    \item \textbf{Oxford RobotCar}~\cite{RobotCarDatasetIJRR}: over 100 traversals of a fixed route through Oxford, across a year under diverse weather, traffic, and construction.
    \item \textbf{Rawseeds}~\cite{fontana2013rawseeds}: Long outdoor and indoor trajectories with structurally similar areas.
    \item \textbf{Habitat-Sim}~\cite{habitat19iccv}: Simulated indoor environments with curated perceptual ambiguities.
    \item \textbf{RELLIS-3D}~\cite{jiang2020rellis3d}: Off-road trajectories with dense vegetation, rough terrain, and aliasing.
    \item \textbf{ROVER}~\cite{schmidt2025rover}: Seasonal outdoor traversals capturing changes in vegetation, terrain, and lighting.
\end{enumerate}

Summary statistics for the benchmark are provided in Table~\ref{tab:dataset_stats_transposed}.

\begin{table*}[t]
\centering
\caption{Localization accuracy of each baseline method, reported at two operating points: the decision threshold $\tau$ corresponding to $\mathcal{L}_{A.O.}@90$ and the threshold yielding $\mathrm{BLA}_{\max}$. For both settings, per-case accuracies (A+P, P.O., A.O.) and the corresponding BLA are shown (see Section~\ref{sec:balanced-localization}).}
\label{tab:balanced_results}
\begin{tabular*}{\textwidth}{@{\extracolsep{\fill}} l cccc cccc}
\hline
\multirow{2}{*}{Method} & \multicolumn{4}{c}{$\mathcal{L}_{A.O.}@90$} & \multicolumn{4}{c}{$\textrm{BLA}_{\textrm{max}}$} \\
\cline{2-5}\cline{6-9}
& $\mathcal{L}_{A+P}$~($\uparrow$) & $\mathcal{L}_{P.O.}$~($\uparrow$) & $\mathcal{L}_{A.O.}$~($\uparrow$) & $\mathrm{BLA}$~($\uparrow$) 
& $\mathcal{L}_{A+P}$~($\uparrow$) & $\mathcal{L}_{P.O.}$~($\uparrow$) & $\mathcal{L}_{A.O.}$~($\uparrow$) & $\mathrm{BLA}$~($\uparrow$) \\

\hline
FAB-MAP~\cite{cummins2009highly} & 0.0 & 0.036 & 0.964 & 0.058 & 0.0 & 0.078 & 0.732 & 0.067 \\
RatSLAM~\cite{milford2008mapping} & 0.0 & 0.081 & 0.438 & 0.057 & 0.0 & 0.081 & 0.438 & 0.057 \\
\midrule
% GM   & 0.020 & 0.399 & 0.690 & 0.189 & 0.196 & 0.692 & 0.157 & 0.280 \\
% SM-Med   & 0.059 & 0.358 & 0.562 & 0.234 & 0.059 & 0.587 & 0.496 & 0.264 \\
% SM-All   & 0.059 & 0.256 & 0.814 & 0.237 & 0.078 & 0.572 & 0.361 & 0.258 \\
% PBU  & 0.020 & 0.399 & 0.500 & 0.170 & 0.157 & 0.681 & 0.35 & 0.337 \\
% \midrule
GM   & 0.020 & 0.402 & 0.577 & 0.178 & 0.078 & 0.479 & 0.423 & 0.256 \\
SM-Med  & 0.039 & 0.311 & 0.778 & \textbf{0.220} & 0.078 & 0.489 & 0.412 & 0.255 \\
SM-All  & 0.020 & 0.229 & 0.784 & 0.164 & 0.039 & 0.388 & 0.469 & 0.200 \\
PBU  & 0.020 & 0.402 & 0.577 & 0.178 & 0.235 & 0.646 & 0.165 & \textbf{0.295} \\
\hline
\end{tabular*}
\end{table*}

\section{Benchmarking Baselines}
\label{sec:benchmarking-baselines}

\subsection{Baseline Systems}
A number of recent works have introduced SLAM-free topological mapping systems (Section~\ref{sec:slam-free-topological-maps}), which differ primarily in how they address the \emph{perceptual aliasing problem}. To study these strategies in depth, we implement three representative approaches that capture their core algorithmic principles.

\paragraph{Sequence Matching (SM).}
Instead of matching a single observation, the system matches short sequences of consecutive observations to improve robustness against perceptual aliasing and viewpoint variation. This temporal context helps reduce false matches caused by transient visual similarities. Representative examples include~\cite{savinov2018semi, meng2020scaling, milford2012seqslam}.  
A candidate shortcut edge $(v_i, v_j)$ is accepted if the aggregated similarity score over a temporal window satisfies
\[
\mathrm{f}\!\left( \mathrm{sim}(z_{v_i-h}, z_{v_j-h}), \ldots, \mathrm{sim}(z_{v_i+h}, z_{v_j+h}) \right) \ge \tau,
\]
where $h$ is the half-window size in frames and $\mathrm{f}\{\cdot\}$ is an aggregation function applied to the similarity scores in the window.  

In our experiments, we evaluate two variants of sequence matching:  
\begin{itemize}
    \item \textbf{SM-Med}: $\mathrm{f}\{\cdot\}$ is the median of the similarity scores, which provides robustness to outliers.  
    \item \textbf{SM-All}: $\mathrm{f}\{\cdot\}$ requires that \emph{all} similarity scores in the window exceed the threshold $\tau$, enforcing stricter consistency across the entire sequence.  
\end{itemize}

\paragraph{Probabilistic belief update (PBU)}  
To maintain temporal consistency, the system maintains a posterior belief $p_t(v)$ over the current node $v$ and updates it at each timestep. Between observations, the belief is propagated using a topological motion model that constrains allowable transitions:
\[
p(v_i \mid v_j) \ \propto \
\begin{cases}
\alpha & \text{if } \mathrm{dist}_{G}(v_i, v_j) \le w_u, \\
\beta  & \text{otherwise},
\end{cases}
\]
where $\mathrm{dist}_{G}(\cdot,\cdot)$ is the hop distance in the graph, $w_u$ is the maximum allowed hop movement per step, and $\alpha,\beta$ are constants controlling transition likelihoods.  

After motion propagation, the belief is updated using the similarity scores as observation likelihoods:
\[
p_t(v) \ \propto\ g\!\left(\mathrm{sim}(z_t, z_v)\right) \cdot \sum_{u \in \mathcal{V}} p(v \mid u) \, p_{t-1}(u),
\]
where $g(\cdot)$ is a likelihood function mapping similarity scores to measurement probabilities.  

This probabilistic filtering suppresses spurious matches that violate motion constraints, improving robustness to perceptual aliasing. Representative examples include~\cite{suomela2024placenav, xu2020probabilistic}.

\paragraph{Greedy Matching (GM)} 
We additionally implement a simple greedy matching baseline, which localizes by selecting the highest-scoring match above a fixed similarity threshold; if no match exceeds the threshold, the localization result remains unchanged.  

In all our baseline implementations, we use the same VPR model MegaLoc~\cite{berton2025megaloc} for fairness, in contrast to prior works that often train environment-specific retrieval models, which may not generalize well to unseen environments. This setup also enables a fair comparison of different topological mapping strategies under identical retrieval performance.

% Read processed CSVs (under data/)
\pgfplotstableread[col sep=comma,trim cells=true]{data/updated_AP.csv}\tblAP
\pgfplotstableread[col sep=comma,trim cells=true]{data/updated_PO.csv}\tblPO
\pgfplotstableread[col sep=comma,trim cells=true]{data/updated_AO.csv}\tblAO
\pgfplotstableread[col sep=comma,trim cells=true]{data/updated_BLA.csv}\tblBLA

\pgfplotsset{
  smmed/.style={mark=none, draw=red!70,   line width=1.1pt},
  gm/.style=   {mark=none, draw=brown!70, line width=1.1pt},
  pbu/.style=  {mark=none, draw=black!60, line width=1.1pt},
  fab/.style=  {mark=none, draw=blue!70,  line width=1.1pt},
  rats/.style= {mark=none, draw=teal!70,  line width=1.1pt},
}

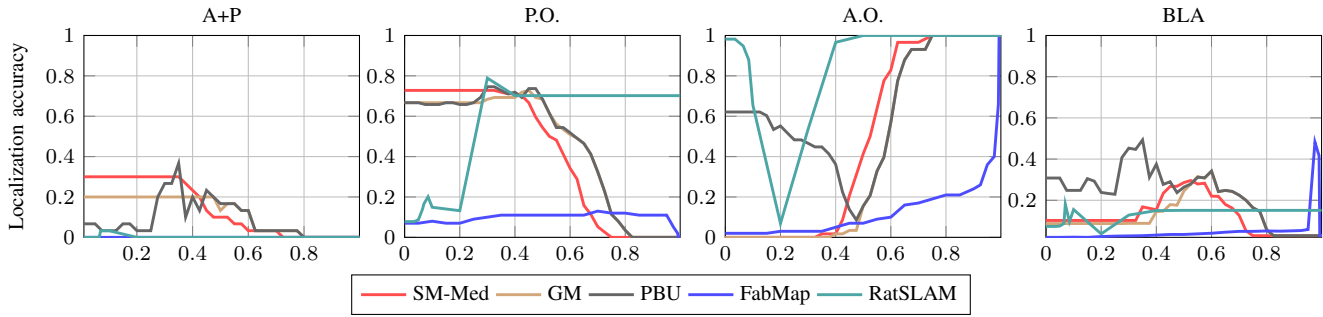
\begin{figure*}[t]
\centering
\begin{tikzpicture}
\begin{groupplot}[
  group style={group size=4 by 1, horizontal sep=6mm}, % tighter gap
  width=0.295\textwidth,          % small bump in width (fits 4x in a row)
  height=0.24\textwidth,        % +20% vs 0.18\textwidth
  grid=both,
  tick label style={font=\footnotesize},
  label style={font=\footnotesize},
  enlarge y limits={abs=0},   % keeps y=0 curve visible
  enlarge x limits={abs=0},   % keeps y=0 curve visible
  ymax=1,                        % same y-scale for all
  title style={font=\footnotesize, yshift=-1ex},
  legend style={font=\footnotesize},
  legend columns=5,
  every axis plot/.append style={no marks}, % lines only
]

% ---- 1) A+P ----
\nextgroupplot[title={A+P}, ylabel={Localization accuracy}, legend to name=sharedlegend]
  \addplot+[smmed] table[x=tau, y={SM-Med}] {\tblAP};
  \addplot+[gm]    table[x=tau, y=GM]       {\tblAP};
  \addplot+[pbu]   table[x=tau, y=PBU]      {\tblAP};
  \addplot+[fab]   table[x=tau, y=FabMap]   {\tblAP};
  \addplot+[rats]  table[x=tau, y=RatSLAM]  {\tblAP};
  \legend{SM-Med,GM,PBU,FabMap,RatSLAM}

% ---- 2) P.O. ----
\nextgroupplot[title={P.O.}]
  \addplot+[forget plot, smmed] table[x=tau, y={SM-Med}] {\tblPO};
  \addplot+[forget plot, gm]    table[x=tau, y=GM]       {\tblPO};
  \addplot+[forget plot, pbu]   table[x=tau, y=PBU]      {\tblPO};
  \addplot+[forget plot, fab]   table[x=tau, y=FabMap]   {\tblPO};
  \addplot+[forget plot, rats]  table[x=tau, y=RatSLAM]  {\tblPO};

% ---- 3) A.O. ----
\nextgroupplot[title={A.O.}]
  \addplot+[forget plot, smmed] table[x=tau, y={SM-Med}] {\tblAO};
  \addplot+[forget plot, gm]    table[x=tau, y=GM]       {\tblAO};
  \addplot+[forget plot, pbu]   table[x=tau, y=PBU]      {\tblAO};
  \addplot+[forget plot, fab]   table[x=tau, y=FabMap]   {\tblAO};
  \addplot+[forget plot, rats]  table[x=tau, y=RatSLAM]  {\tblAO};

% ---- 4) BLA ----
\nextgroupplot[title={BLA}]
  \addplot+[forget plot, smmed] table[x=tau, y={SM-Med}] {\tblBLA};
  \addplot+[forget plot, gm]    table[x=tau, y=GM]       {\tblBLA};
  \addplot+[forget plot, pbu]   table[x=tau, y=PBU]      {\tblBLA};
  \addplot+[forget plot, fab]   table[x=tau, y=FabMap]   {\tblBLA};
  \addplot+[forget plot, rats]  table[x=tau, y=RatSLAM]  {\tblBLA};

\end{groupplot}
\end{tikzpicture}

\vspace{2pt}
\pgfplotslegendfromname{sharedlegend}
\caption{Localization accuracy for A+P, P.O., and A.O. test cases, and BLA, plotted against the decision threshold for all baseline methods. For SM-Med, GM, and PBU the threshold is directly tied to detection scores, while for FabMap and RatSLAM it is defined by more complex scoring functions. Results show that all baseline methods perform poorly in the A+P test case, highlighting the need for benchmarks that explicitly quantify ambiguity and for more sophisticated approaches to resolve perceptual aliasing. In contrast, methods generally perform well in the P.O. test case at low threshold values, since in the absence of perceptual aliasing this setting mainly tests the system’s ability to detect candidates (analogous to recall). For the A.O. test case, methods achieve high accuracy in the high-threshold regime, as the correct outcome is consistently to reject. Finally, the BLA case provides a balanced view across the three scenarios. Overall, these results suggest that current state-of-the-art methods cannot adequately handle scenarios where perceptual aliasing is present.}
\label{fig:locacc_vs_thresh_all}
\vspace{-1em}
\end{figure*}

Finally, we also include specialized strategies that have demonstrated strong performance in large-scale datasets:  

\begin{itemize}
    \item \textbf{FAB-MAP}~\cite{cummins2009highly}: Each observation $Z_k$ is represented as a bag-of-visual-words from SURF features. A Bayesian filter maintains a posterior over location nodes, with likelihoods estimated via a Chow–Liu tree. A motion prior favors adjacent locations, while a Monte Carlo approximated “new-place” hypothesis accounts for novel scenes. Data association selects the most probable location, and top matches are verified with RANSAC.

    \item \textbf{RatSLAM}~\cite{milford2008mapping}: A biologically inspired system where each node (experience) $v_i = \{P^i, V^i, \mathbf{p}_i\}$ stores the pose code $P^i$, local view code $V^i$, and map location $\mathbf{p}_i$. Each edge $e_{ij} \in E$ encodes the odometry-based displacement $\Delta \mathbf{p}_{ij}$ between connected experiences. Upon receiving a new image observation, RatSLAM performs visual template matching to activate corresponding local view cells, applies pose cell filtering with attractor dynamics to integrate visual and self-motion cues, and finally selects the best-matching vertex in $G$ for localization.
\end{itemize}

\subsection{Benchmarking Results}
\label{sec:results}

We evaluated the baseline systems on our ambiguity-quantified dataset, reporting per-case localization accuracies (A+P, P.O., A.O.) as well as the aggregated BLA metric. Following Section~\ref{sec:balanced-localization}, we highlight two complementary operating points: \emph{$\mathcal{L}_{A.O.}@90$}, which anchors evaluation at a safety-critical A.O. reliability level, and \emph{$\mathrm{BLA}{max}$}, which captures the best achievable balance across thresholds. Note that $\mathcal{L}_{A.O.}@90$ is determined by selecting the threshold $\tau$ on a smaller validation set; consequently, the measured $\mathcal{L}_{A.O.}$ on the full dataset may differ. Numerical results are summarized in Table~\ref{tab:balanced_results}, while Fig.~\ref{fig:locacc_vs_thresh_all} plots localization accuracy as a function of threshold for each case type, illustrating how different baselines trade recall against safety.

\paragraph{Overall trends.}
All evaluated methods fall short of the robustness required for real-world deployment. When thresholds are set conservatively so that A.O. cases achieve high rejection accuracy (e.g., 90\% in Table~\ref{tab:balanced_results}), performance on A+P and P.O. cases drops sharply. At even stricter settings ($\mathcal{L}_{A.O.}@99$), every baseline fails completely on A+P and P.O., yielding zero localization accuracy. This reveals a fundamental limitation: current methods address perceptual aliasing only by rejecting more aggressively, but lack robust mechanisms to \emph{disambiguate} visually similar places. As a result, they remain safe only by refusing nearly all revisits. In practice, both failure modes are problematic; for example, a return to the same room may be treated as a new place, fragmenting the map and forcing inefficient routes. These results underscore the lack of robust methods for aliasing and motivate the need for explicitly ambiguity-quantified datasets to expose such trade-offs.

\paragraph{Classical vs. deep-learned VPR.}
Compared with classical approaches such as FAB-MAP~\cite{cummins2009highly} and RatSLAM~\cite{milford2008mapping}, baselines using deep-learned VPR descriptors achieve higher $\mathcal{L}_{A+P}$ and $\mathcal{L}_{P.O.}$ while maintaining a similar level of $\mathcal{L}_{A.O.}$, as reflected in the \emph{$\mathcal{L}_{A.O.}@90$} results in Table~\ref{tab:balanced_results}. This highlights that deep learning–based retrieval offers a stronger foundation for place recognition than hand-engineered pipelines based on feature detection, matching, and geometric verification.

\paragraph{Sequence matching (SM).}
Matching sequences rather than single frames was expected to improve robustness, but in practice the benefit is limited and highly sensitive to the choice of window size $h$ and aggregation function $f(\cdot)$. If $h$ is too short, the method collapses to greedy single-frame matching and fails on ambiguous revisits. If $h$ is too long, test and mapping sequences may diverge due to natural viewpoint changes—for example, the robot may look rightward during testing but straight ahead during mapping—leading the aggregated similarity to fall below threshold even for true revisits. Aggregation with the median is somewhat more robust to outlier mismatches, but this only helps in cases where distractors dominate the top similarity scores; when the true positives are strong, median aggregation can suppress them. As a result, sequence matching baselines do not consistently outperform greedy matching, and their performance curves (see Fig.~\ref{fig:locacc_vs_thresh_all}) show large variance depending on parameter settings.

\paragraph{Probabilistic belief update (PBU).}
Incorporating a topological motion prior via probabilistic filtering yields only modest improvements. The prior is helpful when the system is already well localized and enters an ambiguous segment, as belief can then be reinforced sequentially and spurious matches suppressed. In practice, however, the model assumes consistent hop distances and motion patterns, which rarely hold. For instance, a mapping sequence may contain nodes every 1\,m, while an ambiguous corridor encountered during testing may produce visually similar nodes spaced 2\,m apart. Without precise metric motion information, the prior incorrectly treats these sequences as consistent. This mismatch explains why PBU struggles in practice: the coarse topological prior provides only limited support for disambiguating visually similar places.

\section{Conclusion}

In this paper, we proposed an ambiguity-aware evaluation framework for topological mapping systems. We introduced \emph{localization accuracy} as a surrogate for topological consistency and defined \emph{balanced localization accuracy} (BLA) to aggregate performance across distinct test cases. To support fair and reproducible evaluation, we curated a new dataset with explicit ambiguity quantification.  

Using this dataset, we evaluated both classical approaches and modern systems with deep-learned VPR models. The results yield three main insights: (i) ambiguity-aware evaluation reveals failure modes hidden by conventional loop-closure metrics, as tuning for safe rejection (high $\mathcal{L}_{A.O.}$) causes localization accuracy on revisits (A+P and P.O.) to collapse, showing that current methods cannot truly resolve aliasing; (ii) classical methods remain reliable only for rejecting novel places but fail even on unambiguous revisits, whereas deep-learned retrieval achieves higher revisit accuracy at similar safety levels, reflected in stronger $\mathcal{L}_{A.O.}@90$; and (iii) sequence matching and probabilistic filtering add complexity without consistent gains.  

Overall, these findings underscore the need for more robust disambiguation strategies. Our evaluation metrics and ambiguity-quantified benchmark provide a consistent and reproducible basis for assessing topological mapping systems, fostering further development of reliable and scalable approaches. To ensure reproducibility, we will release all code in a public GitHub repository, which also documents implementation details that cannot be fully included in the paper due to space constraints.

\balance
\bibliographystyle{IEEEtran}  % replaces plainnat

\bibliography{references}

@article{boal2014topological,
  title={Topological simultaneous localization and mapping: a survey},
  author={Boal, Jaime and S{\'a}nchez-Miralles, Alvaro and Arranz, Alvaro},
  journal={Robotica},
  volume={32},
  number={5},
  pages={803--821},
  year={2014},
  publisher={Cambridge University Press}
}

@article{maddern2012cat,
  title={CAT-SLAM: probabilistic localisation and mapping using a continuous appearance-based trajectory},
  author={Maddern, Will and Milford, Michael and Wyeth, Gordon},
  journal={The International Journal of Robotics Research},
  volume={31},
  number={4},
  pages={429--451},
  year={2012},
  publisher={SAGE Publications Sage UK: London, England}
}

@inproceedings{glover2010fab,
  title={FAB-MAP+ RatSLAM: Appearance-based SLAM for multiple times of day},
  author={Glover, Arren J and Maddern, William P and Milford, Michael J and Wyeth, Gordon F},
  booktitle={2010 IEEE international conference on robotics and automation},
  pages={3507--3512},
  year={2010},
  organization={IEEE}
}

@article{lowry2015visual,
  title={Visual place recognition: A survey},
  author={Lowry, Stephanie and S{\"u}nderhauf, Niko and Newman, Paul and Leonard, John J and Cox, David and Corke, Peter and Milford, Michael J},
  journal={ieee transactions on robotics},
  volume={32},
  number={1},
  pages={1--19},
  year={2015},
  publisher={IEEE}
}

@inproceedings{ali2024boq,
  title={BoQ: A place is worth a bag of learnable queries},
  author={Ali-Bey, Amar and Chaib-draa, Brahim and Giguere, Philippe},
  booktitle={Proceedings of the IEEE/CVF Conference on Computer Vision and Pattern Recognition},
  pages={17794--17803},
  year={2024}
}

@inproceedings{shi2019openlorisscene,
    title={Are We Ready for Service Robots? The {OpenLORIS-Scene} Datasets for Lifelong {SLAM}},
    author={Xuesong Shi and Dongjiang Li and Pengpeng Zhao and Qinbin Tian and Yuxin Tian and Qiwei Long and Chunhao Zhu and Jingwei Song and Fei Qiao and Le Song and Yangquan Guo and Zhigang Wang and Yimin Zhang and Baoxing Qin and Wei Yang and Fangshi Wang and Rosa H. M. Chan and Qi She},
    booktitle={2020 International Conference on Robotics and Automation (ICRA)},
    year={2020},
    pages={3139-3145},
}

@article{brooks1991intelligence,
  title={Intelligence without representation},
  author={Brooks, Rodney A},
  journal={Artificial intelligence},
  volume={47},
  number={1-3},
  pages={139--159},
  year={1991},
  publisher={Elsevier}
}

@article{RobotCarDatasetIJRR,
  Author = {Will Maddern and Geoff Pascoe and Chris Linegar and Paul Newman},
  Title = {{1 Year, 1000km: The Oxford RobotCar Dataset}},
  Journal = {The International Journal of Robotics Research (IJRR)},
  Volume = {36},
  Number = {1},
  Pages = {3-15},
  Year = {2017},
  doi = {10.1177/0278364916679498},
  URL =
{http://dx.doi.org/10.1177/0278364916679498},
  eprint =
{http://ijr.sagepub.com/content/early/2016/11/28/0278364916679498.full.pdf+html}}

@incollection{fontana2013rawseeds,
  title={Rawseeds: Building a benchmarking toolkit for autonomous robotics},
  author={Fontana, Giulio and Matteucci, Matteo and Sorrenti, Domenico G},
  booktitle={Methods and experimental techniques in computer engineering},
  pages={55--68},
  year={2013},
  publisher={Springer}
}

@inproceedings{habitat19iccv,
  title     =     {Habitat: {A} {P}latform for {E}mbodied {AI} {R}esearch},
  author    =     {Manolis Savva and Abhishek Kadian and Oleksandr Maksymets and Yili Zhao and Erik Wijmans and Bhavana Jain and Julian Straub and Jia Liu and Vladlen Koltun and Jitendra Malik and Devi Parikh and Dhruv Batra},
  booktitle =     {Proceedings of the IEEE/CVF International Conference on Computer Vision (ICCV)},
  year      =     {2019}
}

@article{schmidt2025rover,
  title={ROVER: A Multiseason Dataset for Visual SLAM},
  author={Schmidt, Fabian and Daubermann, Julian and Mitschke, Marcel and Blessing, Constantin and Meyer, Stephan and Enzweiler, Markus and Valada, Abhinav},
  journal={IEEE Transactions on Robotics},
  volume={41},
  pages={4005--4022},
  year={2025},
  publisher={IEEE}
}

@article{jiang2020rellis3d,
  title={RELLIS-3D Dataset: Data, Benchmarks and Analysis},
  author={Jiang, Peng and Osteen, Philip and Wigness, Maggie and Saripalli, Srikanth},
  journal={arXiv preprint arXiv:2011.12954},
  year={2020}
}

@article{savinov2018semi,
  title={Semi-parametric topological memory for navigation},
  author={Savinov, Nikolay and Dosovitskiy, Alexey and Koltun, Vladlen},
  journal={arXiv preprint arXiv:1803.00653},
  year={2018}
}

@inproceedings{suomela2024placenav,
  title={Placenav: Topological navigation through place recognition},
  author={Suomela, Lauri and Kalliola, Jussi and Edelman, Harry and K{\"a}m{\"a}r{\"a}inen, Joni-Kristian},
  booktitle={2024 IEEE International Conference on Robotics and Automation (ICRA)},
  pages={5205--5213},
  year={2024},
  organization={IEEE}
}

@inproceedings{meng2020scaling,
  title={Scaling local control to large-scale topological navigation},
  author={Meng, Xiangyun and Ratliff, Nathan and Xiang, Yu and Fox, Dieter},
  booktitle={2020 IEEE International Conference on Robotics and Automation (ICRA)},
  pages={672--678},
  year={2020},
  organization={IEEE}
}

@inproceedings{milford2012seqslam,
  title={SeqSLAM: Visual route-based navigation for sunny summer days and stormy winter nights},
  author={Milford, Michael J and Wyeth, Gordon F},
  booktitle={2012 IEEE international conference on robotics and automation},
  pages={1643--1649},
  year={2012},
  organization={IEEE}
}

@article{xu2020probabilistic,
  title={Probabilistic visual place recognition for hierarchical localization},
  author={Xu, Ming and Snderhauf, Niko and Milford, Michael},
  journal={IEEE Robotics and Automation Letters},
  volume={6},
  number={2},
  pages={311--318},
  year={2020},
  publisher={IEEE}
}

@article{milford2008mapping,
  title={Mapping a suburb with a single camera using a biologically inspired SLAM system},
  author={Milford, Michael J and Wyeth, Gordon F},
  journal={IEEE Transactions on Robotics},
  volume={24},
  number={5},
  pages={1038--1053},
  year={2008},
  publisher={IEEE}
}

@article{cummins2009highly,
  title={Highly scalable appearance-only SLAM-FAB-MAP 2.0},
  author={Cummins, M and Newman, P},
  journal={Robotics: Science and Systems V},
  year={2009},
  publisher={Robotics: Science and Systems Foundation}
}

@inproceedings{blochliger2018topomap,
  author    = {Fabian Bl{\"o}chliger and Marius Fehr and Marcin Dymczyk and Thomas Schneider and Roland Siegwart},
  title     = {Topomap: Topological Mapping and Navigation Based on Visual SLAM Maps},
  booktitle = {Proc. IEEE International Conference on Robotics and Automation (ICRA)},
  pages     = {3818--3825},
  year      = {2018},
  doi       = {10.1109/ICRA.2018.8460641}
}

@inproceedings{oleynikova2018sparse,
  author    = {Helen Oleynikova and Zachary Taylor and Roland Siegwart and Juan Nieto},
  title     = {Sparse 3D Topological Graphs for Micro-Aerial Vehicle Planning},
  booktitle = {Proc. IEEE/RSJ International Conference on Intelligent Robots and Systems (IROS)},
  year      = {2018}
}

@inproceedings{konolige2011hybrid,
  author    = {Kurt Konolige and Eitan Marder{-}Eppstein and Bhaskara Marthi},
  title     = {Navigation in Hybrid Metric--Topological Maps},
  booktitle = {Proc. IEEE International Conference on Robotics and Automation (ICRA)},
  pages     = {3041--3047},
  year      = {2011}
}

@inproceedings{friedman2007vrf,
  author    = {Stephen Friedman and Hanna Pasula and Dieter Fox},
  title     = {Voronoi Random Fields: Extracting the Topological Structure of Indoor Environments via Place Labeling},
  booktitle = {Proc. Int. Joint Conf. on Artificial Intelligence (IJCAI)},
  year      = {2007}
}

@inproceedings{beeson2005evg,
  author    = {Patrick Beeson and Nicholas K. Jong and Benjamin Kuipers},
  title     = {Towards Autonomous Topological Place Detection Using the Extended Voronoi Graph},
  booktitle = {Proc. IEEE International Conference on Robotics and Automation (ICRA)},
  pages     = {4373--4379},
  year      = {2005},
  doi       = {10.1109/ROBOT.2005.1570793}
}

@article{wang2025genie,
  title={GeNIE: A Generalizable Navigation System for In-the-Wild Environments},
  author={Wang, Jiaming and Liu, Diwen and Chen, Jizhuo and Da, Jiaxuan and Qian, Nuowen and Man, Tram Minh and Soh, Harold},
  journal={arXiv preprint arXiv:2506.17960},
  year={2025}
}

@inproceedings{sridhar2024nomad,
  title={Nomad: Goal masked diffusion policies for navigation and exploration},
  author={Sridhar, Ajay and Shah, Dhruv and Glossop, Catherine and Levine, Sergey},
  booktitle={2024 IEEE International Conference on Robotics and Automation (ICRA)},
  pages={63--70},
  year={2024},
  organization={IEEE}
}

@inproceedings{berton2025megaloc,
  title={Megaloc: One retrieval to place them all},
  author={Berton, Gabriele and Masone, Carlo},
  booktitle={Proceedings of the Computer Vision and Pattern Recognition Conference},
  pages={2861--2867},
  year={2025}
}

@inproceedings{berton2023eigenplaces,
  title={Eigenplaces: Training viewpoint robust models for visual place recognition},
  author={Berton, Gabriele and Trivigno, Gabriele and Caputo, Barbara and Masone, Carlo},
  booktitle={Proceedings of the IEEE/CVF International Conference on Computer Vision},
  pages={11080--11090},
  year={2023}
}

@inproceedings{wang2024polo,
  author={Wang, Jiaming and Soh, Harold},
  title={Probable Object Location ({POLo}) Score Estimation for Efficient Object Goal Navigation},
  booktitle={2024 IEEE International Conference on Robotics and Automation (ICRA)},
  year={2024},
  pages={5221--5227},
  publisher={IEEE},
  url={https://ieeexplore.ieee.org/document/10610671/}
}

\end{document}